\documentclass{article}
\usepackage{spconf,amsmath,graphicx,bm}
\usepackage{hyperref}
\usepackage[table]{xcolor}
\usepackage{multirow}
\usepackage{url}
\usepackage{makecell}
\usepackage[normalem]{ulem}
\usepackage{caption}

\newif\ifreview
\reviewtrue

\ifreview
\newcommand{\rv}[2][]{%
    \if\relax\detokenize{#1}\relax%
        \textcolor{blue}{#2}%
    \else%
        \textcolor{gray}{\sout{#1}}\textcolor{red}{#2}%
    \fi%
}
\newcommand{\todo}[1]{\noindent\textcolor{orange}{To-Do: {#1}}}
\newcommand{\comment}[2][]{
    \if\relax\detokenize{#1}\relax%
        \noindent\textcolor{magenta}{Comment: {#2}}%
    \else%
        \noindent\textcolor{magenta}{Comment from #1: {#2}}%
    \fi%
}
\else
\newcommand{\rv}[2][]{#2}
\newcommand{\todo}[1]{\empty}
\newcommand{\comment}[2][]{\empty}
\fi

\newcommand{\best}[1]{\cellcolor{gray}\textcolor{white}{\textbf{{#1}}}}

\newcommand{\red}[1]{\textcolor{red}{\textbf{{#1}}}}

\title{Mixup-based Deep Metric Learning Approaches \\ for Incomplete Supervision \\ \small{\red{(Extended Version)}}}

\name{\begin{tabular}{c}Luiz H. Buris$^1$\red{*}, Daniel C. G. Pedronette$^2$, Joao P. Papa$^3$, Jurandy Almeida$^4$,\\ Gustavo Carneiro$^5$, Fabio A. Faria$^{1,5}$\red{*}\end{tabular}\thanks{The authors are grateful to the S{\~a}o Paulo Research Foundation (FAPESP) grants 	2021/01870-5, 2018/23908-1, 2014/12236-1, 2017/25908-6, and 2019/07665-4, to the National Council for Scientific and Technological Development (CNPq) grants 308529/2021-9 and 314868/2020-8, and to NVIDIA for donating a Titan V GPU used in the experiments. \red{\textbf{*L.H.Buris and F.A.Faria contributed equally to this work (co-first authorship).}}}}
\address{$^1$Institute of Science and Technology, Universidade Federal de S\~{a}o Paulo - Brazil\\
$^2$Depart. of Statistics, Applied Math. and Computing, S\~ao Paulo State University - Brazil\\
$^3$Department of Computing, S\~ao Paulo State University - Brazil\\
$^4$Department of Computing, Federal University of S\~ao Carlos - Brazil\\
$^5$Australian Institute for Machine Learning, The University of Adelaide - Australia}

\begin{document}
\maketitle

\begin{abstract}
Deep learning architectures have achieved promising results in different areas (e.g., medicine, agriculture, and security). However, using those powerful techniques in many real applications becomes challenging due to the large labeled collections required during training. Several works have pursued solutions to overcome it by proposing strategies that can learn more for less, e.g., weakly and semi-supervised learning approaches. 
As these approaches do not usually address memorization and sensitivity to adversarial examples, this paper presents three deep metric learning approaches combined with Mixup for incomplete-supervision scenarios. We show that some state-of-the-art approaches in metric learning might not work well in such scenarios.  Moreover, the proposed approaches outperform most of them in different datasets.

\end{abstract}

\section{Introduction}
\label{s.introduction}

Deep Learning~(DL) architectures, in particular Convolutional Neural Networks (CNNs), have been intensively studied due to their great success in different application domains (e.g., action recognition~\cite{SIBGRAPI_2020_Santos}, biometric recognition~\cite{dl_biometric_2018}, object segmentation~\cite{dl_semantica_segmentation_2018}, medical image analysis~\cite{andrade_2018}, and crop yield estimation in agriculture~\cite{aono_plosone2021}). Despite the excellent outcomes, their training requires large labeled collections that may not be feasible in real applications~\cite{BucciLT20}.

Many works have strived to learn the target distribution using reduced labeled sets and, at the same time, take advantage of large amounts of unlabeled data~\cite{SIBGRAPI_2021_Silva}. These solutions primarily use the weakly supervised learning framework, which is related to different modalities of data supervision~\cite{wsl_survey}. Diverse alternatives rely on semi-supervised learning, which leverages unlabeled data to guide models during the training process (e.g., consistent regularization~\cite{Hu-2020-SimPLE}, pseudo-labeling~\cite{pseudolabel-icml2013}, and label propagation~\cite{IscenTAC_CVPR19}).

Weakly supervised learning approaches can be broadly characterized by three kinds of data supervisions: (1) \emph{Incomplete} supervision, in which the amount of labeled data is insufficient to train a good model (reduced training scenario); (2) \emph{Inexact} supervision addresses the lack of information and accuracy about the provided data, and (3) \emph{Inaccurate} supervision, where the given labels are not always correct as labeling errors (noisy labels) may be present~\cite{wsl_survey}. 


Some tasks, such as learning manifolds (aka metric learning), are extremely challenging in weakly supervised scenarios. Usually, such approaches consider the samples' neighborhoods (some distance function is applied) to either map them onto new embeddings or understand their structure. Deep Metric Learning (DML) aims to learn such embeddings using deep learning architectures, with many promising works, such as Nearest Neighbour Gaussian Kernel (NNGK)~\cite{NNGK_ICIPP2018}, SoftTriple~\cite{SoftTriple_iccv2019}, ProxyAnchor~\cite{ProxyAnchor_cvpr2020}, and Supervised Contrastive (SunCon)~\cite{SupCon_nips2020}, to cite a few.

However, DML approaches do not usually address memorization and are prone to adversarial attacks. 
Mixup~\cite{zhang2018mixup} is a simple learning principle that tries to overcome the aforementioned drawbacks. 
Still, to the best of our knowledge, it has never been applied to the context of metric learning for weakly supervised classification. 
Mixup is considered an input regularization method and, despite being straightforward, provides promising results~\cite{SantosCS:22}. 

In this paper, we propose three new approaches in the context of DML. We are particularly interested in NNGK due to its robustness and simplicity. As such, we introduce variants that take advantage of Mixup to cope with metric learning in incomplete supervision scenarios.




In summary, the contributions of this paper are threefold: (a) to evaluate different state-of-the-art (SOTA) metric learning approaches in incomplete supervision scenarios; (b) to introduce three Mixup-based metric learning approaches for incomplete supervision scenarios; and (c) to improve NNGK approach in image classification tasks.

\section{Related Works}
\label{s.related_works}
Metric learning approaches commonly use either contrastive, triplet, or neighborhood analysis to build loss functions. Contrastive loss is the most straightforward and intuitive approach, for it aims to minimize the distance between pairs of samples from the same class and separate them otherwise. Also, it uses a margin strategy to prevent the mapping function maps all data to the same point in the embedding space, making distances between samples to be equal to zero~\cite{Contrastive_cvpr2006}. Triplet loss, inspired in the context of nearest-neighbor classification, ensures that an anchor (positive class) is closer to all positive samples than to any other negative data point~\cite{triplet_cnn_cvpr2015}. A margin value is applied between positive and negative pairs. 

Neighborhood Component Analysis (NCA)~\cite{goldberger2004neighbourhood} estimates a Mahalanobis distance measure to be used in the well-known $k$-Nearest neighbors classification problem. It optimizes the leave-one-out performance on training data through a differentiable cost function that uses neighbors' stochastic (soft) assignments in the transformed space.

Nearest Neighbour Gaussian Kernel (NNGK)~\cite{NNGK_ICIPP2018} combines metric learning and classification concepts using a particular loss function for training CNNs, projecting the original embeddings to a Gaussian kernel space as an approximate nearest neighbor search problem. SoftTriple~\cite{SoftTriple_iccv2019} loss learns the embeddings without the sampling phase. Qian et al.~\cite{SoftTriple_iccv2019} showed that minimizing the normalized SoftMax loss is equivalent to optimizing a smoothed triplet loss. In this loss, the number of triplets is linear to the number of original examples.  

Proxy-Anchor loss~\cite{ProxyAnchor_cvpr2020} takes advantage of pair-based and proxy-based losses and overcomes their limitations. As a proxy represents a subset of training data, proxy-based losses can reduce the training complexity and convergence time once the number of representatives is significantly smaller. Proxy-Anchor loss uses each proxy as an anchor and associates it to all data from a given batch during training. SupCon~\cite{SupCon_nips2020} extends the contrastive self-supervised framework to a fully-supervised setting, effectively leveraging label information. Points that belong to the same class are likely to be mapped to the same cluster in the embedded space; samples from different categories are kept apart.

In spite of all the advances, most existing approaches still suffer from a lack of memorization and robustness to adversarial attacks. This paper aims to fill such a gap. For this, we combine DML with Mixup and introduce three new approaches for incomplete-supervision scenarios.
To the best of our knowledge, we are the first to investigate the combination of DML and Mixup for weakly supervised classification.

\section{Proposed Approaches}
\label{s.proposed_approaches}

\begin{figure*}[!htb]
\centering
\includegraphics[width=1\textwidth]{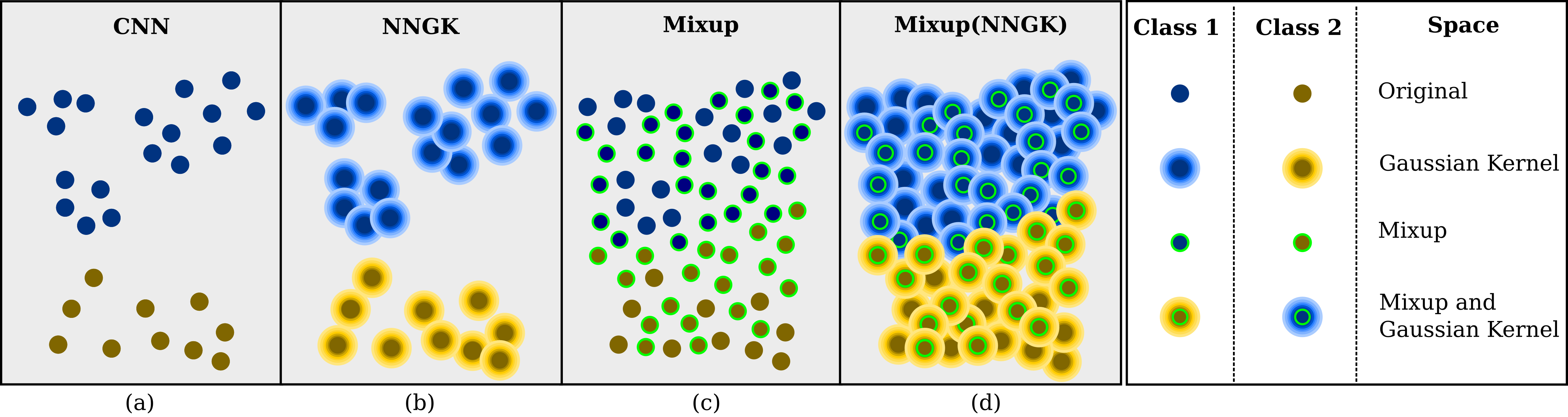}
\caption{Different embeddings defined by each approach: (a) samples in the feature space defined by a pre-trained CNN, (b) same samples projected onto a Gaussian kernel, (c) samples in the feature space of the pre-trained CNN together with the new samples created by Mixup, and (d) samples in the feature space by the combination of NNGK and Mixup. Notice that, in this paper, there are these four possible kinds of feature spaces, therefore the three proposed approaches based on Mixup (\textit{MbDML}) are some combination of the existing feature spaces.}
\label{fig:spaces}
\end{figure*}

This section presents the three new approaches proposed in this paper. For the sake of explanation, we also provide a brief introduction to NNGK and Mixup techniques\footnote{For explanation purposes, we defined all the loss functions in the manuscript for a single image.}.

\subsection{Nearest Neighbour Gaussian Kernels}
\label{ss.nngk}

Nearest Neighbour Gaussian Kernel (NNGK)~\cite{NNGK_ICIPP2018} adopts a pre-trained CNN that projects its original feature embeddings to a Gaussian kernel space and updates its weights by minimizing the NCA loss during the training process. 

Let ${\cal X} = \{(\bm{x}_1,\nu(\bm{x}_1)),(\bm{x}_2,\nu(\bm{x}_2)),\ldots,(\bm{x}_m,\nu(\bm{x}_m))\}$ be a training dataset such that $\bm{x}_i\in\mathcal{R}^d$ stands for an embedding in some $d$-dimensional space and $\nu(\bm{x}_i)\in{\cal Y}$ denotes its label, where ${\cal Y}=\{\omega_1,\omega_2,\ldots,\omega_c\}$. Additionally, let $\mathcal{C}=\{\bm{c}_1,\bm{c}_2,\dots,\bm{c}_n\}$ be a set of $n$ Gaussian kernel centers such that $\phi(\bm{x}_i)\in{\cal C}$ assigns every sample $\bm{x}_i\in{\cal X}$ to its nearest-with-the-same-label center. Equation~\ref{Eq:nngk} computes a Gaussian classifier that outcomes the probability of a given dataset sample $\bm{x}$ belonging to some class label $\nu(\bm{x})\in\mathcal{Y}$:
\small{
\begin{equation}
     P(\nu(\bm{x})=\omega_i|\bm{x}) =  \frac{\displaystyle\sum_{\bm{z}\in\mathcal{N}(\bm{x})\wedge\nu(\bm{z})=\omega_i}{w_{\phi(\bm{z})}\frac{\|\bm{z}-\phi(\bm{z})\|^2}{2\sigma^2}  }} {\displaystyle\sum_{\bm{k}\in \mathcal{N}(\bm{x})}{w_{\phi(\bm{k})}  \frac{\|\bm{k}-\phi(\bm{k})\|^2}{2\sigma^2} }}, 
    \label{Eq:nngk}
\end{equation}
}

\noindent where $\mathcal{N}(\bm{x})$ stands for the set of approximate nearest neighbours of $\bm{x}$ and $w_{\phi(\bm{z})}$ denotes the weight for center $\phi(\bm{z})$, which is learned end-to-end with the network weights. The loss function, based on the NCA loss, is defined as follows:
\begin{equation}
{\cal L}_{NNGK} = \sum_{\omega_i\in{\cal Y}} -\ln(P(\nu(\bm{x})=\omega_i|\bm{x})).
\end{equation}

\subsection{\textit{Mixup: Beyond Empirical Risk Minimization}}
\label{sec:mixup}

Mixup~\cite{zhang2018mixup} uses a simple learning principle to overcome problems on the CNN training process related to memorization and sensitivity to adversarial examples. It generates weighted combinations of random image pairs from the training data and their respective one-hot labels. The method fills the region between classes in the problem space by making decision boundaries smoother and more regular through virtual data augmentation on the training set. The choice of data to be combined runs at random, as well as the selection of $\lambda\in[0,1]$ which is the mixing factor

Let ${\cal I}=\{(\bm{I}_1,y_1),(\bm{I}_2,y_2),\ldots,(\bm{I}_M,y_M)\}$ be a dataset of images $\bm{I}_i$ with their corresponding one-hot labels $y_i$. Given two image-label pairs $(\bm{I}_i,y_i)$ and $\bm{I}_j,y_j)$ such that $y_i\neq y_j$, we can define:
\begin{equation}
 \hat{\bm{I}} = \lambda \times \bm{I}_i + (1 -\lambda )\times \bm{I}_j\text{\ \ \ \ \ and}   
 \label{Eq:I}
\end{equation}
\begin{equation}
    \hat{y} = \lambda \times y_i + (1 -\lambda )\times y_j,
\end{equation}

\noindent where $\hat{\bm{I}}$ and $\hat{y}$ stand for the ``mixed"\ image and its one-hot soft multi-label, respectively.


Let $CE(\bm{I},y) = p(\bm{I}_i)\times \log p(y_i)$ be the cross-entropy function for a given image $\bm{I}\in{\cal I}$, where $p(\bm{I}_i)$ denotes the probability vectors extracted from the output of the CNNs and $p(y_i)$ denotes its corresponding label vector. The Mixup loss function is computed as follows:
\begin{equation}
  {\cal L}_{Mixup} =  \lambda \times CE(\hat{\bm{I}},{y_i}) + (1-\lambda) \times CE(\hat{\bm{I}},{y_j}), 
  \label{Eq:mixup}
\end{equation}

\noindent where $y_i\neq y_j$.


\subsection{Mixup-based Deep Metric Learning Approaches (\textit{MbDML})}
\label{sec:proposal}
 
This section describes the three proposed approaches based on Mixup, illustrated in Figure~\ref{fig:spaces}.

\subsubsection{MbDML 1: $NNGK+Mixup$}

The first approach computes a sum between the two original loss functions from NNGK and Mixup techniques: 
\begin{equation}
  {\cal L}_1 =  {\cal L}_{NNGK} + {\cal L}_{Mixup}.
  \label{Eq:nngk+mixup}
\end{equation}
 
\subsubsection{MbDML 2:  $Mixup(NNGK)$} 

The second approach uses the probability vectors computed by a Gaussian classifier (Equation~\ref{Eq:nngk}) as input instead of the original probability vector (softmax function) employed in $CE$. Moreover, it also applies the interpolation factor $\lambda$: 
%
%
%
\begin{equation}
\begin{aligned}
{\cal L}_2 = {} & \lambda \times \left( P(\nu(\bm{x})=\omega_i|\bm{x})\times \log p(y) \right)  \\
& + (1-\lambda) \times \left( P(\nu(\bm{x})=\omega_j|\bm{x})\times \log p(y) \right).
\end{aligned}
\label{Eq:mixup(dml)}
\end{equation}



\subsubsection{MbDML 3: $Mixup(NNGK) + NNGK$}

This last approach combines two more complex loss functions, i.e., ${\cal L}_{2}$ and ${\cal L}_{NNGK}$, using a scaling factor $\alpha$:
%
%
\begin{equation}
  {\cal L}_3 = \alpha \times {\cal L}_2 + {\cal L}_{NNGK}.
  \label{Eq:mixup(dml)+nngk}
\end{equation}


\section{Experiments and Discussion}
\vspace{-1mm}
In this section, we provide details about the experimental setup and report the obtained results.
A rigorous and extensive experimental evaluation was conducted on four different datasets. 
The proposed approaches were compared to state-of-the-art methods for incomplete and complete supervision scenarios.

\subsection{Datasets}
\vspace{-1mm}
We adopt four different datasets in the incomplete supervision experiments, e.g., Cifar10, Cifar100, Flowers17, and MNIST. Moreover, we partitioned the training sets into $10$ folds to simulate a reduced training scenario. In the complete supervision experiments, where the entire training set is used for training purposes, three datasets (Flowers102, Cars196, and LeafsnapField) were adopted respecting their original setup (training and test sets). However, they have been added to the original training set when validation sets exist. Table~\ref{tab:dataset} describes the datasets in a more detailed fashion.

\begin{table}[!htb]
     \caption{Description of each dataset used in the experiments.}
    \centering \resizebox{8.5cm}{!}{
    \begin{tabular}{lcccc}  \hline
    \textbf{Dataset} & \textbf{Classes} & \textbf{Resolution} & \textbf{Train} & \textbf{Test} \\\hline
    \multicolumn{5}{c}{  Incomplete Supervision Scenario}\\\hline
    Cifar10~\cite{cifar_2009} & 10 & 32$\times$32 & 50k & 10k \\
    Cifar100~\cite{cifar_2009} & 100 & 32$\times$32 & 50k & 10k  \\
    Flowers17~\cite{flowers_2008} &17 & 256$\times$256 &1,190 & 170 \\
    MNIST~\cite{mnist_2010} & 10 & 28$\times$28 & 50k & 10k  \\\hline
    \multicolumn{5}{c}{ Complete Supervision Scenario} \\\hline
    Flowers102~\cite{flowers_2008} & 120 & 256$\times$256 & 2040 & 5,532 \\
    Cars196~\cite{cars_2013} & 196 & 256$\times$256 & 8,144 & 8,041 \\
    LeafsnapField~\cite{leafsnap_2012} & 184 & 256$\times$256 & 4,514 & 885 \\\hline
    \end{tabular}
    }
   \label{tab:dataset}
\end{table}

\begin{table*}[!htb]
\caption{\red{Mean accuracies ($\%$) and standard deviation ($\pm$) over ten runs using 10\% of the training set. Similar and the most accurate results are highlighted.}}
\centering
\resizebox{.8\textwidth}{!}{
\begin{tabular}{clcccc}
 \multicolumn{2}{c}{ \multirow{2}{*}{\textbf{Approaches}} } & \multicolumn{4}{c}{ \textbf{ Datasets} } \\\cline{3-6}
&& \textbf{Cifar10} & \textbf{Cifar100} & \textbf{MNIST} & \textbf{Flowers17}  \\\hline
\multirow{5}{*}{\rotatebox[origin=c]{0}{\makecell{\textbf{Losses}}}}
& Contrastive~\cite{Contrastive_cvpr2006} & 67.02 $\pm$ 1.37 &26.69	 $\pm$ 0.53  & 98.51 $\pm$ 0.27 & 46.47 $\pm$ 5.15\\ 
&Proxy-Anchor~\cite{ProxyAnchor_cvpr2020} & \textcolor{red}{69.06} $\pm$ 	1.08  &  24.97 $\pm$ 	1.32  & \textcolor{red}{98.90} $\pm$ 0.11  &  \textcolor{red}{82.57}	$\pm$ 2.81 \\ 
&SoftTriple~\cite{SoftTriple_iccv2019}  & 66.14 $\pm$ 	0.61 & 26.44 $\pm$ 	1.41 & 98.75 $\pm$ 0.11 &  64.31	$\pm$ 5.44 \\ 
&SupCon~\cite{SupCon_nips2020}  & 62.53 $\pm$ 1.37 & 10.85 $\pm$ 0.61 & 98.75  $\pm$ 0.27 & 61.81 $\pm$ 9.38\\
&Triplet~\cite{triplet_cnn_cvpr2015} & 62.55 $\pm$ 0.73 &  \textcolor{red}{32.70} $\pm$ 6.05 & 98.57 $\pm$ 0.21 & 54.21 $\pm$ 4.96\\ \hline 	
\multirow{2}{*}{\rotatebox[origin=c]{0}{\makecell{\textbf{Originals}}}}&Mixup~\cite{zhang2018mixup} & 75.20 $\pm$ 8.88 & \textcolor{blue}{38.80} $\pm$ 4.66 & 97.90 $\pm$ 0.32 & 58.46 $\pm$ 3.12\\
&NNGK~\cite{NNGK_ICIPP2018} & \textcolor{blue}{77.07} $\pm$ 0.34 & 38.26 $\pm$ 0.69 & \textcolor{blue}{97.96} $\pm$ 0.19 & \textcolor{blue}{72.42} $\pm$ 3.44\\ \hline 
\multirow{4}{*}{\rotatebox[origin=c]{0}{\makecell{\textbf{Ours}}}}&{\textit{MbDML} 1} & \best{78.95 $\pm$ 0.32} & 46.31 $\pm$ 0.27 & \best{98.53 $\pm$ 0.07} & 80.60 $\pm$ 1.88\\
& {\textit{MbDML} 2} & 77.89 $\pm$ 0.40 & \best{49.43 $\pm$ 0.39} & 97.47 $\pm$ 0.20 & \best{82.62 $\pm$ 1.27}\\ 
& {\textit{MbDML} 3} & 78.63 $\pm$ 0.46 & 47.83 $\pm$ 0.46 & 98.21 $\pm$ 0.11 & 82.36 $\pm$ 1.09\\ \hline \hline 
\textbf{Relative}&\textcolor{red}{Ours $\times$ Losses} & 14.3& 51.2 & -0.3 & 0.0\\
\textbf{Gain}& \textcolor{blue}{Ours $\times$ Originals}  & 2.4 & 27.4 & 0.6 & 14.1\\
\hline 
\end{tabular}
}
\label{tab:comparison_approaches}
\end{table*}

\subsection{Experimental Setup}

Source codes provided by the authors of NNGK\footnote{\tiny{\url{https://github.com/LukeDitria/OpenGAN}} (As of February, 2022)} and Mixup\footnote{\tiny{\url{https://github.com/facebookresearch/mixup-cifar10}} (As of February, 2022)} techniques are used in the manuscript. We considered NNGK with two centers, i.e., $|\mathcal{C}|=\{100,200\}$, and two different values for $\sigma=\{5,10\}$. Mixup default parameters from the original paper are used here. Concerning the loss-based approaches, i.e., Contrastive, Proxy-Anchor, SoftTriple, SupCon, and Triplet, the source codes available at PyTorch Metric Learning lib~\footnote{\tiny{\url{https://kevinmusgrave.github.io/pytorch-metric-learning/}} (As of February, 2022)} were used as a baseline using their default parameters. Last but not least, the three proposed approaches were implemented upon the source codes of the combined approaches (NNGK and Mixup). We adopted a pre-trained ResNet50~\cite{resnet} with a batch size of $32$ images trained by $200$ epochs implemented on the PyTorch library~\cite{pytorch_NEURIPS2019_9015} as the backbone for all techniques. For only one experiment with the original Mixup (Flowers17 dataset), Resnet50 model was exchanged for another Resnet50 capable of learning with images of dimensions $256 \times 256$.  


\subsection{Incomplete Supervision Scenarios}

The experimental evaluation faces a challenging scenario, for we are dealing with a learning task in a reduced-labeled dataset. In the incomplete supervision approach, we assume the existence of a data collection composed of labeled and a much larger unlabeled subset. The experiments consider we have only $10\%$ of the entire datasets figuring some label. The remaining samples have no information about it. As this experiment aims to analyze the behavior of metric learning approaches in scaled-down training scenarios, the unlabeled data set is not used during training. Therefore, we understand that using semi-supervised approaches is unfair and will be a natural advance of this paper.

Table~\ref{tab:comparison_approaches} presents the classification results concerning the three proposed approaches compared to NNGK, Mixup, and six state-of-the-art methods (Contrastive, Proxy-Anchor, SoftTriple, SupCon, and Triplet) over four image datasets (Cifar10, Cifar100, Flowers17, and MNIST). SOTA metric learning approaches based on losses do not perform well in the incomplete supervision scenarios for almost all datasets, except for Proxy-Anchor, which figured a slight advantage in the MNIST and good performance in the Flowers17 dataset. Furthermore, all proposed approaches achieved the best classification results for Cifar10 and Cifar100 datasets compared to the baseline approaches. Last but not least, \textit{MbDML} 1 yielded similar classification results to the best baseline approach (Proxy-Anchor) in the MNIST dataset. Also, it is slightly more accurate in the Flowers17 dataset.

\red{As a matter of interest, Tables~\ref{fig:curves} and~\ref{fig:tsne} show the learning curves and t-SNE visualization of the first round of the performed experiments for incomplete supervision scenarios.}


\subsection{Complete Supervision Scenarios}

This section compares NNGK, initially suggested for complete supervision scenarios, against the three proposed approaches (\textit{MbDML} 1, \textit{MbDML} 2, and \textit{MbDML} 3). Table~\ref{tab:our-NNGK} presents the classification results for the three datasets (Flowers102, Cars196, and LeafsnapField) used in the NNGK original paper~\cite{NNGK_ICIPP2018}. All  \textit{MbDML} approaches achieved more accurate results than NNGK for all datasets, with relative gains between $3.13\%$ and $6.91\%$. Furthermore, this fact shows that the combination of NNGK and Mixup methods can be a promising solution, increasing accuracy in image classification tasks.

\begin{table}[!th]
\centering
\caption{{Classification results for complete supervision scenarios. The best results  are highlighted.}}
\label{tabOurs}
\begin{tabular}{lccc} 
\multirow{2}{*}{\textbf{Approaches}} & \multicolumn{3}{c}{ \textbf{ Datasets} } \\\cline{2-4}
 & \textbf{Flowers102} & \textbf{Cars196} & \textbf{LeafsnapField}\\ \hline
NNGK~\cite{NNGK_ICIPP2018} & \textcolor{red}{87.80} & \textcolor{red}{82.73} & \textcolor{red}{90.40}\\ \hline
MbDML 1 & 89.61 & 86.69 & 92.74 \\ \hline
MbDML 2 & 90.48 & 88.31 & 92.63 \\ \hline
MbDML 3 & \best{90.55} & \best{88.45} & \best{93.08} \\ \hline\hline
\textbf{Gain} &\textbf{3.13}   &\textbf{6.91} & \textbf{2.96} \\ \hline
\end{tabular}
 \label{tab:our-NNGK}
\end{table}

\begin{table*}[ht!]
    \centering
    \begin{tabular}{cc}
    \multicolumn{2}{c}{\textbf{Cifar10}} \\ 
       \includegraphics[scale=.35]{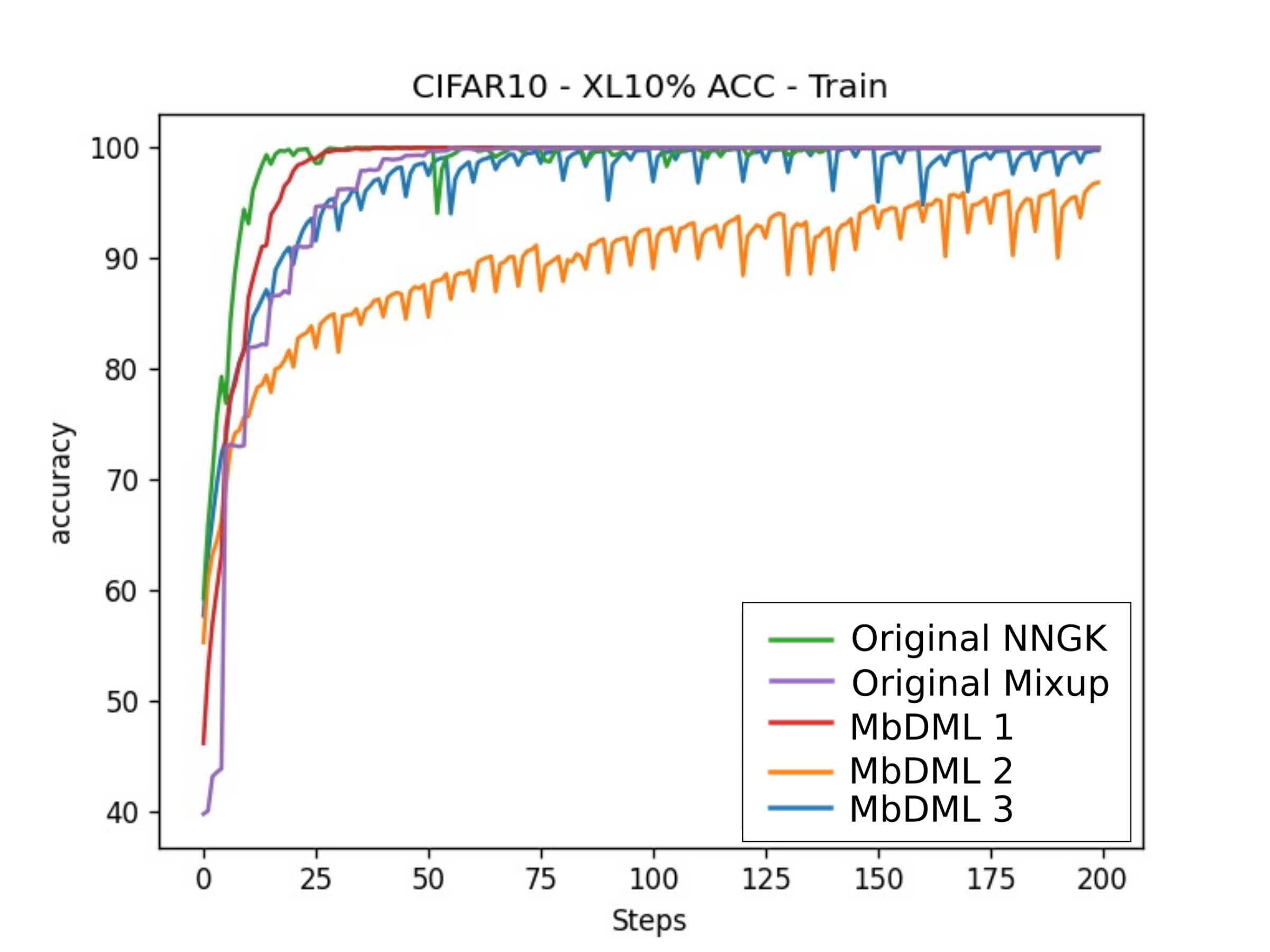} & \includegraphics[scale=.35]{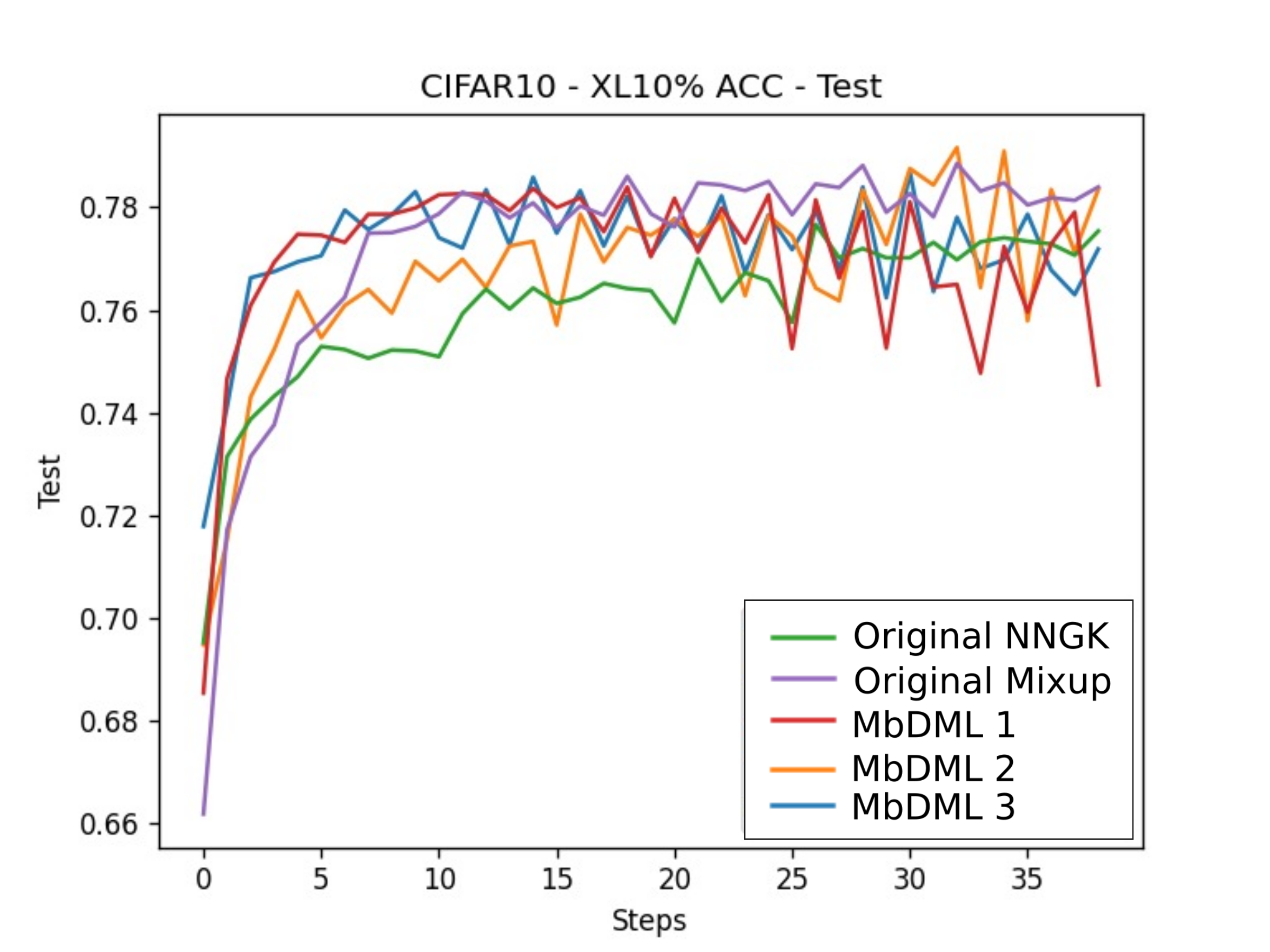}\\ 

      \multicolumn{2}{c}{\textbf{Cifar100}} \\
      \includegraphics[scale=.35]{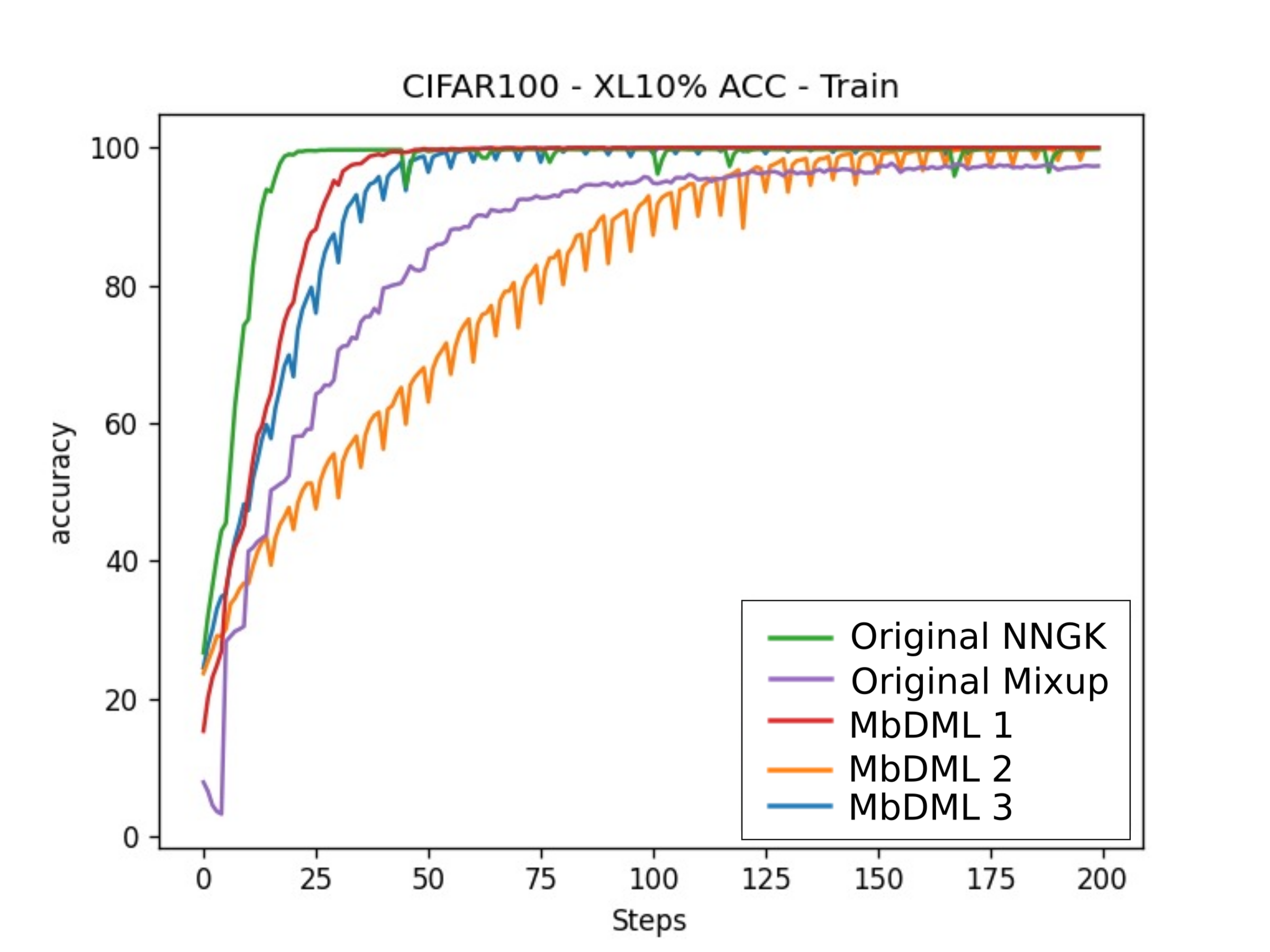} & \includegraphics[scale=.35]{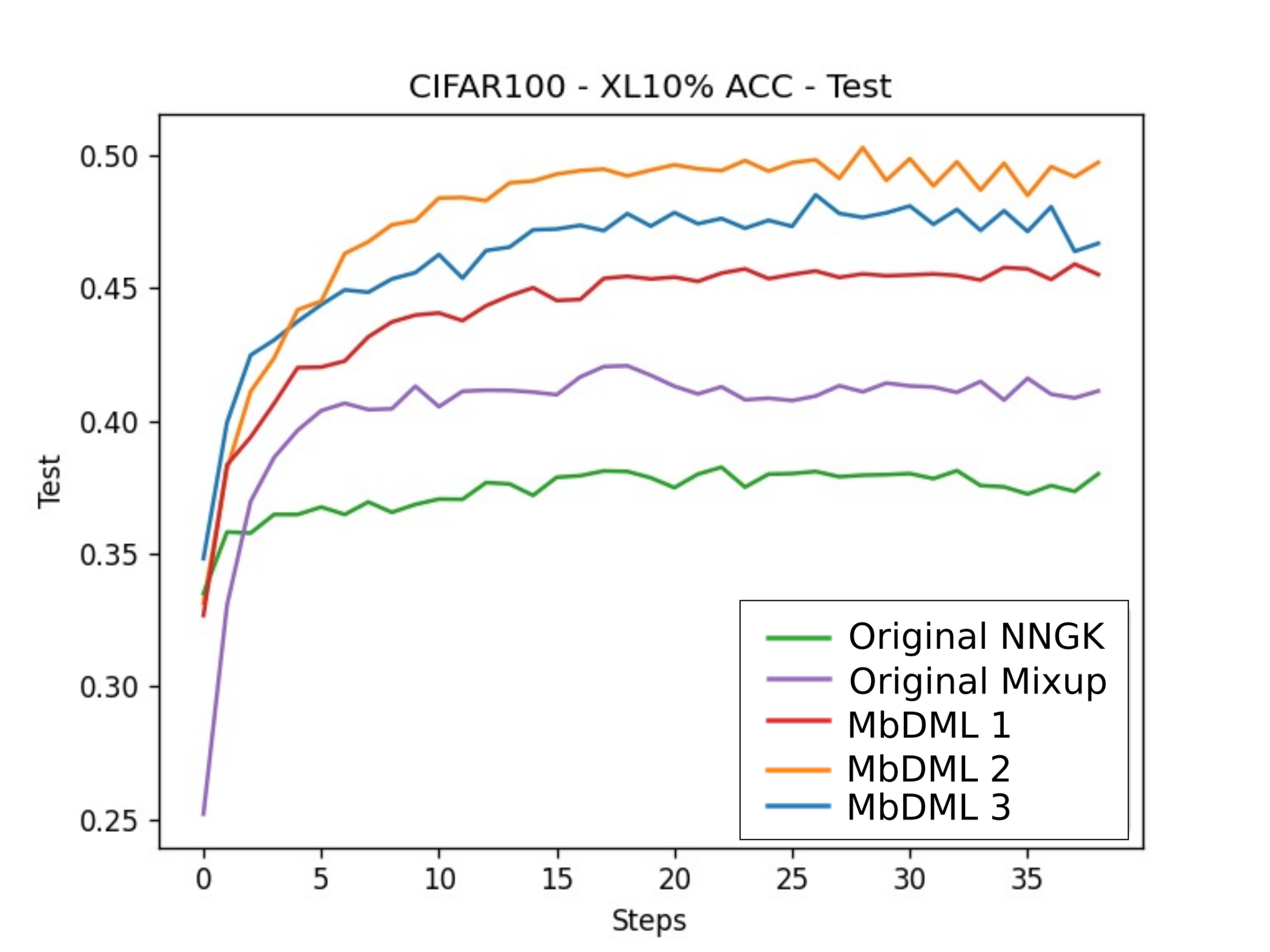} \\

        \multicolumn{2}{c}{\textbf{MNIST}} \\
       \includegraphics[scale=.35]{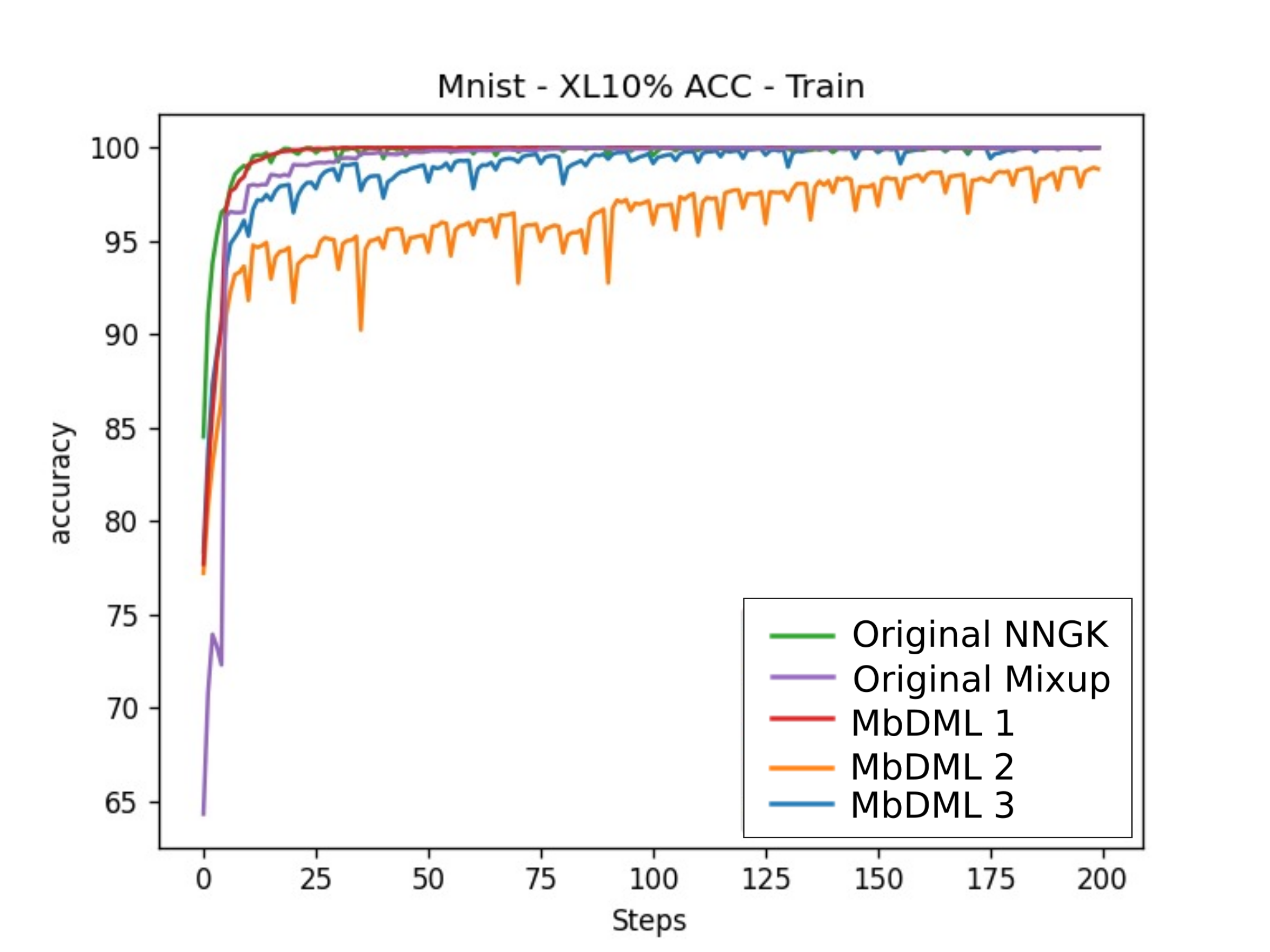} & \includegraphics[scale=.35]{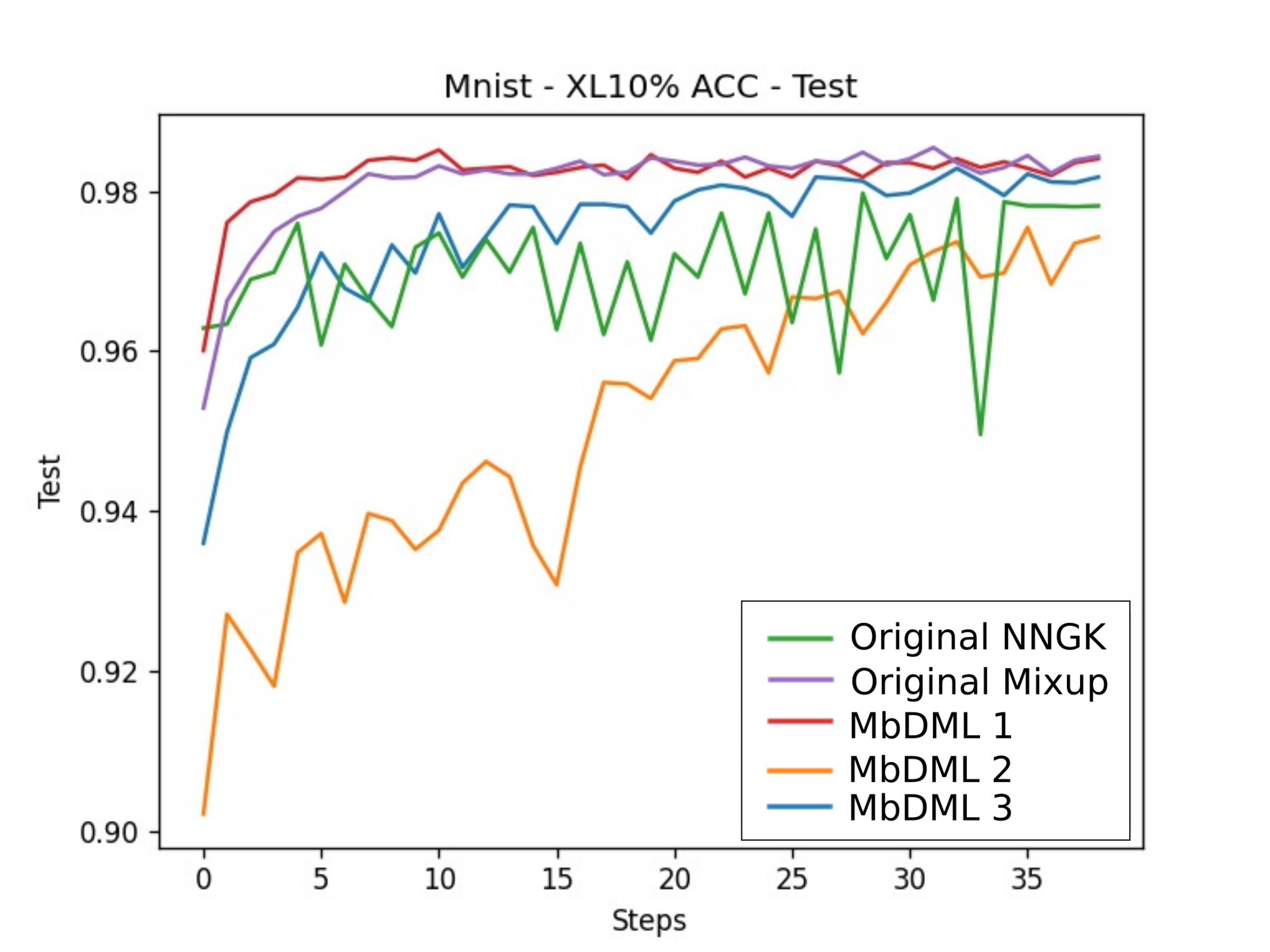} \\
        \multicolumn{2}{c}{\textbf{Flowers17}} \\
       \includegraphics[scale=.35]{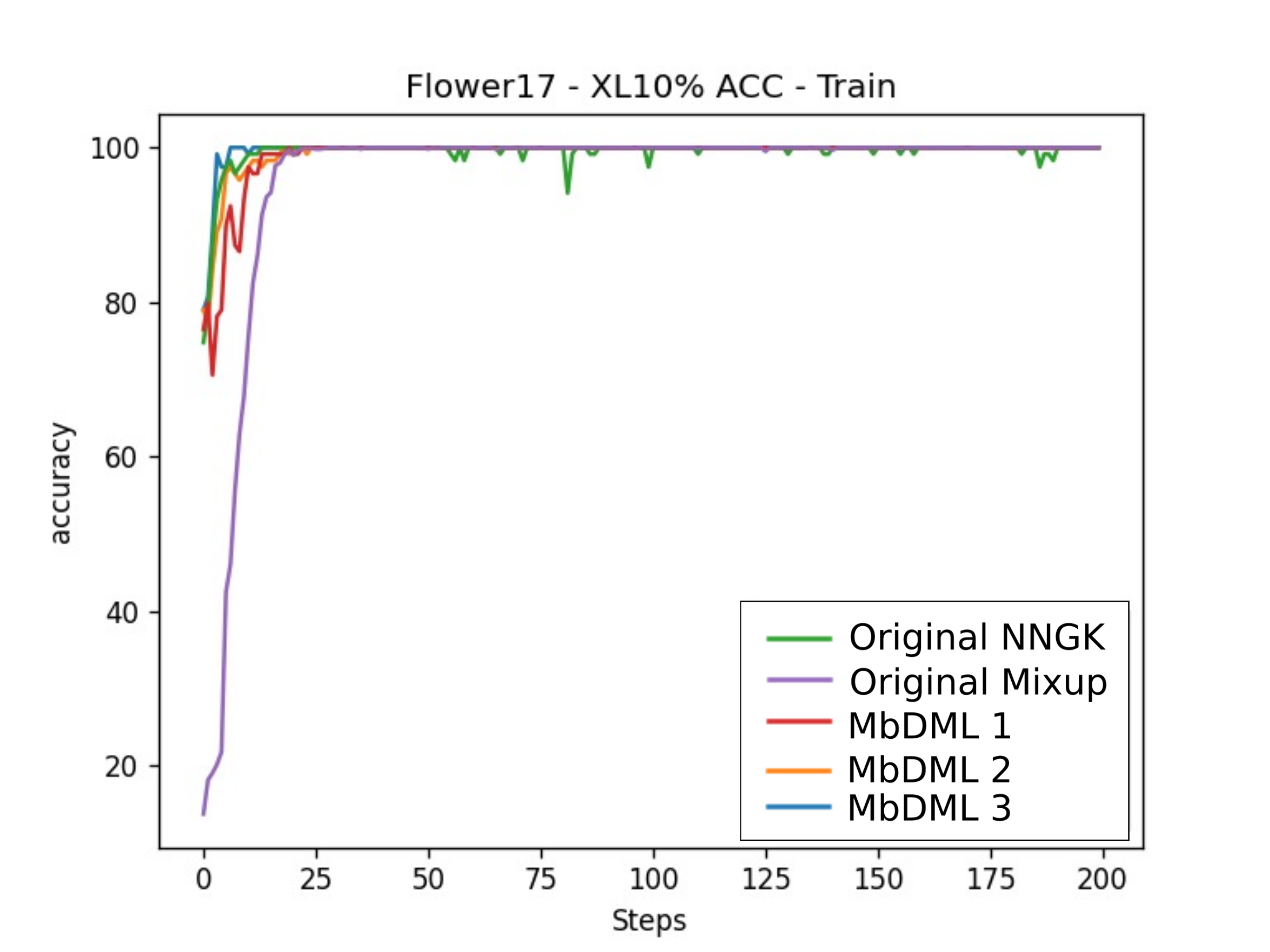} & \includegraphics[scale=.35]{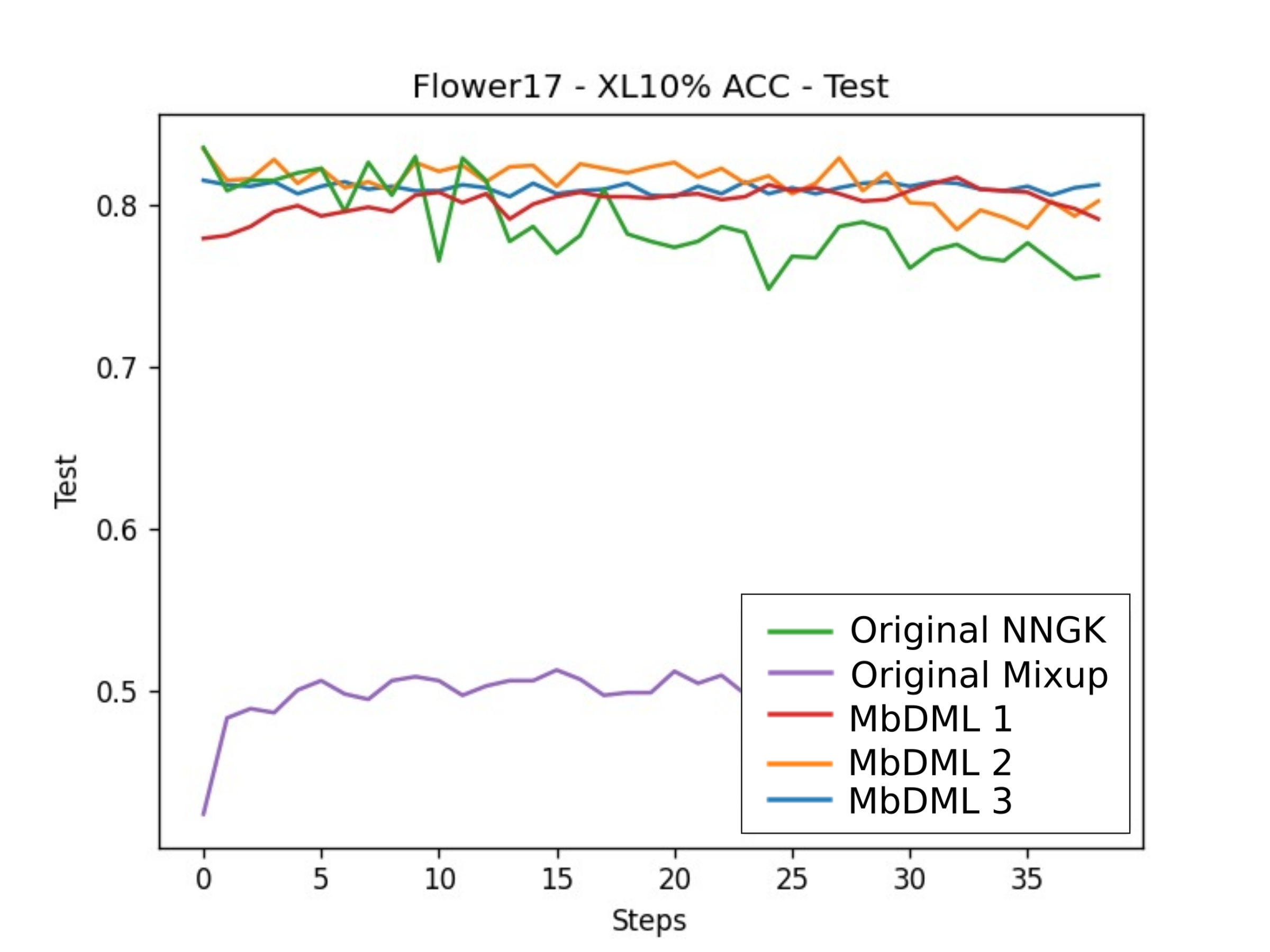} \\
        (a) Training Set.    &    (b) Test Set. \\
    \end{tabular}
    \caption{\red{Accuracy curves among the originals  approaches (NNGK and Mixup) and the $MbDML$ approaches proposed for incomplete supervision scenarios at the first round of the performed experiments on training and test sets.}}
    \label{fig:curves}
\end{table*}

\begin{table*}[ht!]
    \centering
    \begin{tabular}{cc}
    \multicolumn{2}{c}{\textbf{Cifar10}} \\
      \includegraphics[width=.35\textwidth]{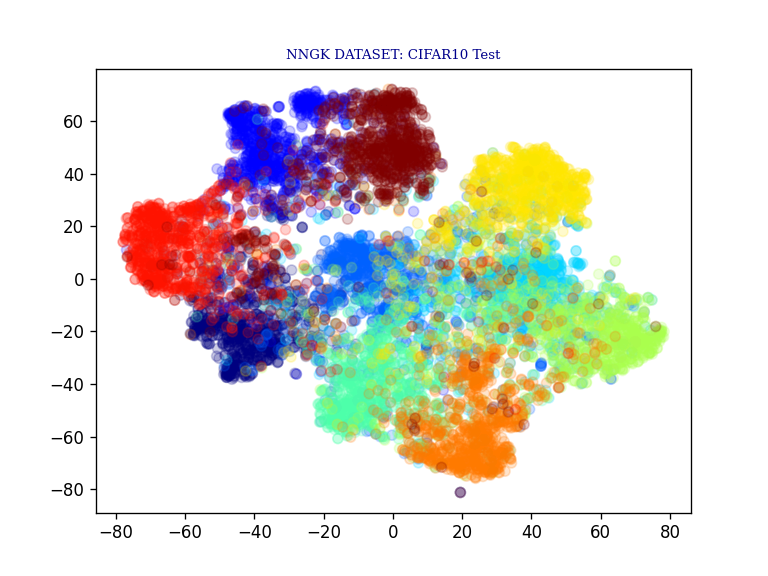}   & \includegraphics[width=.35\textwidth]{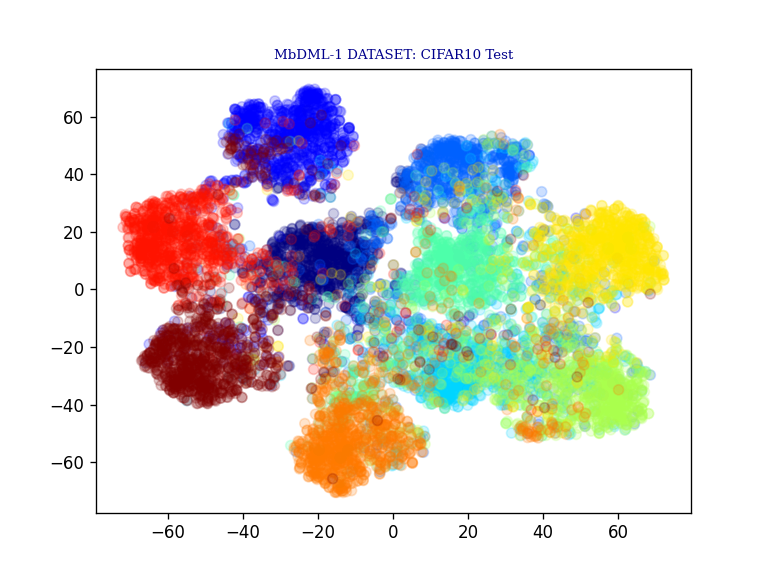} \\ 
      (a) NNGK. & (b) \textit{MbDML} 1. \\ 
    \multicolumn{2}{c}{\textbf{Cifar100}} \\
  \includegraphics[width=.35\textwidth]{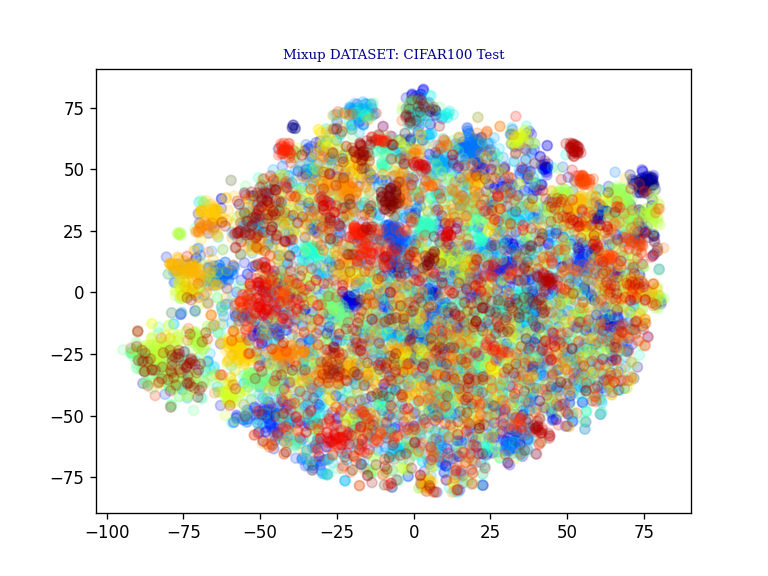}   & \includegraphics[width=.35\textwidth]{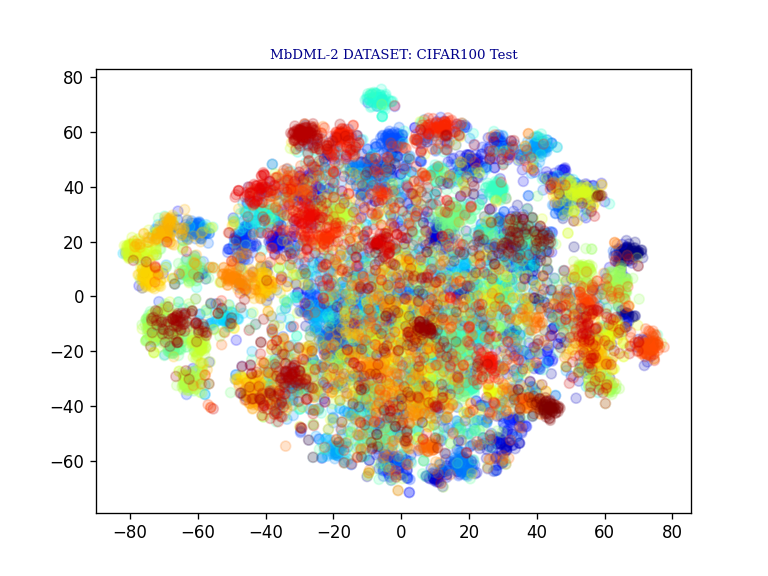} \\ 
  (c) Mixup. & (d) \textit{MbDML} 2. \\ 
        \multicolumn{2}{c}{\textbf{MNIST}} \\
        \includegraphics[width=.35\textwidth]{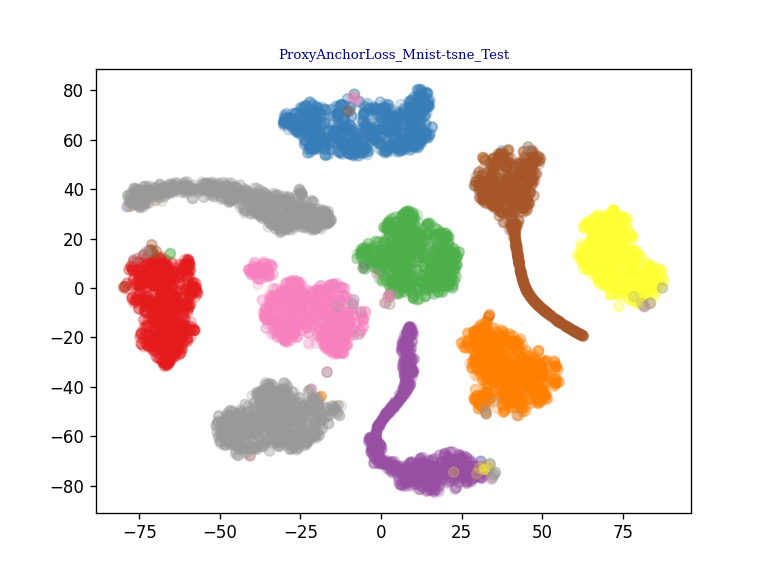}   & \includegraphics[width=.35\textwidth]{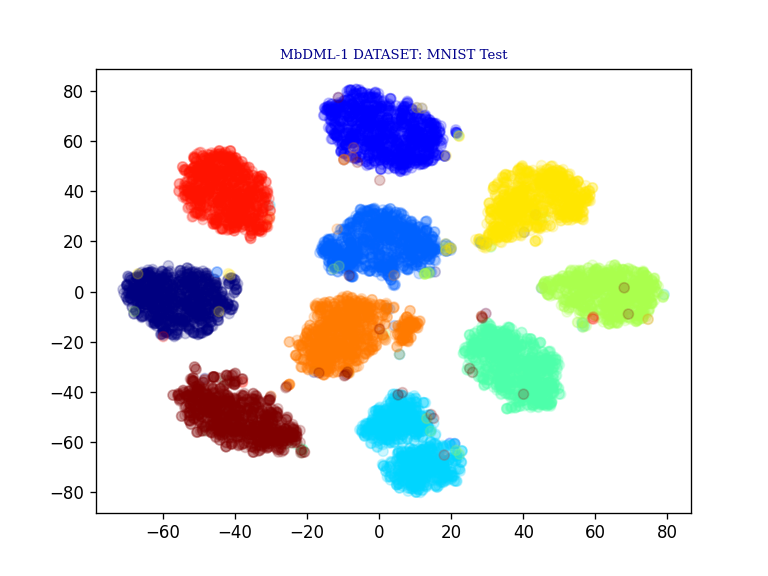} \\   (e) Proxy-Anchor. & (f) \textit{MbDML} 1. \\ 
             \multicolumn{2}{c}{\textbf{Flowers17}} \\ 
             \includegraphics[width=.35\textwidth]{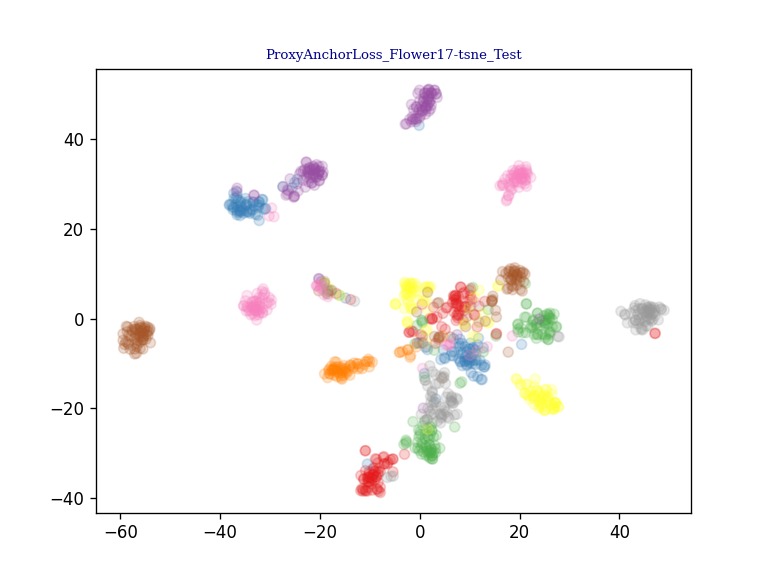} & \includegraphics[width=.35\textwidth]{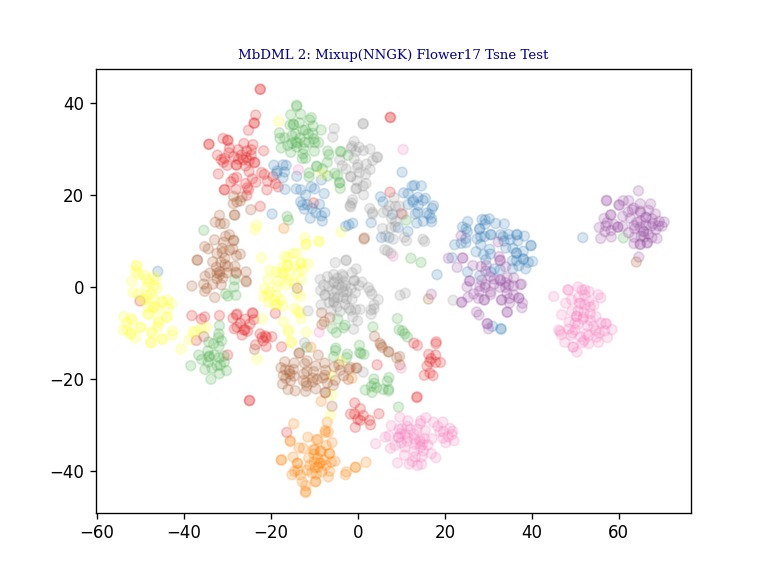} \\ 
         (g) Proxy-Anchor.  & (h) \textit{MbDML} 2. \\
    \end{tabular}
    \caption{\red{t-SNE visualization of the best baseline and \textit{MbDML} approaches (see Table~\ref{tab:comparison_approaches}) for each dataset  adopted in this work at the first round of the performed experiments.}}
    \label{fig:tsne}
\end{table*}
\clearpage

\section{Conclusion}

This paper proposes three DML approaches based on Mixup to improve classification results in incomplete supervision scenarios. These new approaches were validated on four different datasets (Cifar10, Cifar100, Flowers17, and MNIST), outperform all the SOTA approaches in two datasets, and achieved similar results in the other two. Furthermore, the proposed methods achieved more accurate results than NNGK in three datasets  (Flowers102, Cars196, and LeafsnapField) for the complete supervision scenario. As future work, we intend to evaluate the proposed approaches on other datasets and also with different CNN architectures, as well as to extend the \textit{MbDML} approaches to semi-supervised learning tasks.

\small{
\bibliographystyle{IEEEbib}
\bibliography{refs}
}
\end{document}